\pdfoutput=1
\documentclass[11pt]{article}

\usepackage[]{emnlp2021}

\usepackage{times}
\usepackage{latexsym}

\usepackage[T1]{fontenc}
\usepackage[utf8]{inputenc}

\usepackage{microtype}

\usepackage{amsmath,amssymb,amsfonts}
\usepackage{url,enumitem}
\usepackage{booktabs}
\usepackage{xspace}
\usepackage{graphicx}
\usepackage{subcaption}
\usepackage{comment}
\usepackage[normalem]{ulem}
\newcommand{\Sref}[1]{\S\ref{#1}}

\title{Influence Tuning: Demoting Spurious Correlations via\\Instance Attribution and Instance-Driven Updates}

\author{
  Xiaochuang Han \quad \quad \quad \quad Yulia Tsvetkov\\
  Paul G.~Allen School of Computer Science \& Engineering, University of Washington \\
        {\tt \{xhan77, yuliats\}@cs.washington.edu}
}

\begin{document}
\maketitle
\begin{abstract}
Among the most critical limitations of deep learning NLP models are their lack of interpretability, and their reliance on spurious correlations. 
Prior work proposed various approaches to interpreting the black-box models to unveil the spurious correlations, 
but the research was primarily used in human-computer interaction scenarios. It still remains underexplored whether or how such model interpretations can be used to automatically ``unlearn'' confounding features. 
In this work, we propose \emph{influence tuning}---a procedure that leverages model interpretations to update the model parameters towards a plausible interpretation (rather than an interpretation that relies on spurious patterns in the data) in addition to learning to predict the task labels. 
We show that in a controlled setup, influence tuning can help deconfounding the model from spurious patterns in data, significantly outperforming baseline methods that use adversarial training.\footnote{This work was done at Carnegie Mellon University. Code is available at \url{https://github.com/xhan77/influence-tuning}.}
\end{abstract}

\section{Introduction}
\label{sec:intro}
Despite the huge success of contemporary deep learning models 
and various applications that they power, critical limitations persist. 
Among the most harmful issues are their lack of interpretability  \citep{Lipton2018TheMO,guidotti2018survey},
and the tendency to learn spurious correlations, 
in addition to the true signals of the task \citep{leino2019feature,sagawa2020investigation}. 
Both of these lead to corrosive outcomes, 
from reduced performance on datasets in which the confounds no longer hold \citep{jia2017adversarial,Gururangan2018AnnotationAI,glockner2018breaking,McCoy2019RightFT,kumar2019topics,clark2019don}, 
to pernicious biases in model decisions \citep{sun2019mitigating,blodgett2020language,field-etal-2021-survey}, 
and to overall reduced trust in technology \citep{Ribeiro2016WhySI,ehsan2019automated}. 

Consequently, multiple approaches have been proposed to alleviate 
the issues of the growing inscrutability and brittleness of the models.  
Two prominent approaches to interpretability in NLP models are 
(1) \emph{feature attribution}---identifying important tokens in the input span, 
e.g.~via saliency maps \citep{Li2016VisualizingAU, Ribeiro2016WhySI}; 
and (2) \emph{instance attribution}---explaining the model decisions 
as a function of influential training data \citep{Koh2017UnderstandingBP, Han2020ExplainingBB, Pruthi2020EstimatingTD}. 
Both lines of research aim to help users build trust in the model 
by showing the rationale behind the model decision.

Approaches to demoting the influence of spurious confounds in the data include 
(1) model-based approaches to learn confound-invariant representations, e.g., adversarial training  \citep{pryzant2018deconfounded,elazar2018adversarial,kumar2019topics}; 
(2) data-based approaches to balance the training data, e.g., counterfactual data augmentation    \citep{zmigrod-etal-2019-counterfactual,Kaushik2020LearningTD}; 
(3) optimization approaches to account for worst-case scenarios, e.g., distributionally robust optimization  \citep{Sagawa2019DistributionallyRN}; 
and (4) post-processing approaches, such as model ensembling \citep{clark2019don}.

The issues of interpretability and robust generalization are not unrelated. 
Interpretations can facilitate the discovery of the 
model's reliance on frequent spurious patterns.
For example, in natural language inference models an over-reliance on lexical signals can be revealed via feature attribution \citep{Gururangan2018AnnotationAI}, via instance attribution \citep{Han2020ExplainingBB}, or through a combination of thereof \citep{Pezeshkpour2021CombiningFA}.
In this work, we investigate a closer interaction between 
interpretability and model robustness. 

Our research hypothesis is that interpretations that discover confounds 
can be incorporated at training time, to proactively guide the model 
towards avoiding the confounds and improving generalization. 
Our method relies on instance attribution interpretation methods that determine the \emph{influence} of training data on the model's decisions (\Sref{sec:instance_attribution}).
We show how this influence can help discover the model's reliance on some spurious patterns, first in an illustrative task (\Sref{sec:motivation}), and then more generally in our proposed framework \emph{influence tuning}. Influence tuning aims to demote the spurious patterns by guiding the model
to produce plausible interpretations 
via instance attribution (\Sref{sec:influence_tuning}).
We evaluate our approach on two datasets in a controlled setup (\Sref{sec:experimental_setup}, \Sref{sec:results}). Our experiments show that the proposed influence tuning method outperforms the baselines that use adversarial training \citep{ganin2016domain, pryzant2018deconfounded}. We conclude with a discussion of a potentially broader impact of influence tuning on various NLP tasks. 

\section{Interpretation via Instance Attribution}
\label{sec:instance_attribution}
Our primary goal is to use model interpretations for deconfounding the model during training. 
We focus on instance attribution approaches since these interpretations may help capture higher-level attributes in addition to token- and phrase-level lexical features, e.g., span overlaps, the length of text, etc.
In this section, we review the family of instance attribution methods. 

Many NLP models share a same general formula for their decision process during testing: $\hat{y} = f(x_{\text{test}}; \theta)$, where $x_{\text{test}}$ is the test input tokens and $\theta$ is the parameters of the trained model. While feature attribution methods like saliency maps \citep{Simonyan2013DeepIC, Li2016VisualizingAU} focus on interpreting an NLP model's decision by the importance of each individual tokens within $x_{\text{test}}$, instance attribution  methods often look at the influence of $\theta$ on the decision, which is further influenced by the training examples the model uses during the training phase.

\paragraph{Influence functions}
\citet{Koh2017UnderstandingBP} propose influence functions (IF) for ML models, following the vision from robust statistics. IF first approximates how upweighting a particular training example $z_{\text{train}} = (x_{\text{train}}, y_{\text{train}})$ in the training set $\{(x_1, y_1), \ldots, (x_n, y_n)\}$ by an infinitesimal $\epsilon_{\text{train}}$ would change the learned model parameters $\theta$:
\begin{align*}
    \frac{d\theta}{d\epsilon_{\text{train}}} = - H_{\theta}^{-1} \nabla_{\theta}{\mathcal{L}(z_{\text{train}}, \theta)} \text{,}
\end{align*}
where $H_{\theta} = \frac{1}{n} \sum_{i=1}^{n} \nabla_{\theta}^{2}{\mathcal{L}(z_i, \theta)}$ is the Hessian of the model. We can then use the chain rule to measure how this change in the model parameters would in turn affect the loss of the probing input:\footnote{A probing input can be either obtained during test time or selected from the training set.}
\begin{align*}
    \frac{d\mathcal{L}(z_{\text{probe}}, \theta)}{d\epsilon_{\text{train}}} = \nabla_{\theta}{\mathcal{L}(z_{\text{probe}}, \theta)} \cdot \frac{d\theta}{d\epsilon_{\text{train}}}
\end{align*}
The final influence of a train example to a probing example is defined as: $\mathcal{I}(z_{\text{train}}, z_{\text{probe}}) = - \frac{d\mathcal{L}(z_{\text{probe}}, \theta)}{d\epsilon_{\text{train}}}$. That is, a training example is influential to a probing example if upweighting it in the training set would make the model more likely to make a correct decision over the probing example.\footnote{More details about IF and its applications in NLP can be found in \citet{Koh2017UnderstandingBP} and \citet{Han2020ExplainingBB}.}

\paragraph{Gradient product}
Computing the inverse Hessian $H_{\theta}^{-1}$ in the IF is expensive and requires further approximations if the model is non-convex.\footnote{\citet{basu2020influence} also points out IF can be less accurate when used with deep neural networks.} \citet{Pruthi2020EstimatingTD} tackle the problem from a different perspective and arrive at a similar, but a first-order solution:\footnote{Here, we are presenting the equally weighted TracInCP variant from \citet{Pruthi2020EstimatingTD}.}
\begin{align*}
    \mathcal{I}(z_{\text{train}}, z_{\text{probe}}) = \sum_{i=1}^{k} &\nabla_\theta \mathcal{L}(z_{\text{train}}, \theta_i)\\
    \cdot &\nabla_\theta \mathcal{L}(z_{\text{probe}}, \theta_i) 
    \text{,}
\end{align*}
where $\theta_i$ is the checkpoint of the model at each training epoch. The intuition behind this method is to approximate the total reduction in the probing loss $\mathcal{L}(z_{\text{probe}}, \theta)$ during the training process when the training example $z_{\text{train}}$ is used. Compared to IF, this gradient product method essentially drops the inverse Hessian term and reduces the problem to the dot product between the gradient of the training loss and the gradient of the probing loss.

\paragraph{Gradient cosine}
One potential problem of IF and gradient product is being dominated by some outlier training examples, where the norm of their training gradients is significantly larger than the rest of examples. This would lead the method to identify the same set of outlier training examples being influential to a large number of different probing examples. \citet{Barshan2020RelatIFIE} points out this variance-lacking problem of IF and proposes a simple modification: substituting the dot product operation with cosine similarity, normalizing by the norm of the training gradients. Following the same intuition, we modify and further simplify the gradient product method:
\begin{align*}
    \mathcal{I}(z_{\text{trn}}, z_{\text{prb}}) = \frac{\nabla_\theta \mathcal{L}(z_{\text{trn}}, \theta) \cdot \nabla_\theta \mathcal{L}(z_{\text{prb}}, \theta)}{\|\nabla_\theta \mathcal{L}(z_{\text{trn}}, \theta)\|~\|\nabla_\theta \mathcal{L}(z_{\text{prb}}, \theta)\|}
\end{align*}
We use this latter influence definition for the instance attribution interpretation method throughout this work.

\section{A~Toy~Example:~Predicting~Text~Length}
\label{sec:motivation}
Now let's imagine a simple synthetic task where an NLP model like BERT \citep{Devlin2019BERTPO} is trained for binary text classification. 
Class $0$ contains random short sentences with a length sampled from $\mathcal{N}(\mu_{\text{short}}, \sigma^2)$; 
class $1$ contains random long sentences with a length sampled from $\mathcal{N}(\mu_{\text{long}}, \sigma^2)$. 
Our classification task is to predict the text length. 

However, there are confounds in this data. 
For every sentence, we insert a confounding token at the beginning of the sentence. 
Most of the time (e.g., 90\%), token \texttt{A} would co-occur with a short sentence and \texttt{B} would co-occur with a long sentence; for the remaining sentences, this co-occurrence is flipped. 

Our goal is to finetune the classifier so that it learns to predict the class labels ($0$ or $1$) 
using, as intended, the text length information, 
instead of overfitting to the confounding tokens \texttt{A} or \texttt{B}. 
We refer to the text length as a \emph{core attribute}, and to the confounding prefix tokens as a \emph{spurious attribute}.\footnote{This toy task is inspired by the numeric toy dataset in \citet{sagawa2020investigation}.}

Finetuning the classifier on our synthetic task yields an 100\% accuracy on the training set (overfitting). We are more interested in interpreting what information the model relies on to make classification decisions. 
This drives us to apply the instance attribution interpretation methods.

To interpret each classification decision via instance attribution, we randomly sample a few examples within the training set\footnote{In our experiments we choose probing examples from the training set, but it can be a held-out set as well.} as our probing examples~$z_{\text{prb}}$. 
We calculate the influence of each training example $z_{\text{trn}}$ on $z_{\text{prb}}$ using the gradient cosine method (\Sref{sec:instance_attribution}). 
Our expectation is that the $z_{\text{trn}}$ instances that have the same core attribute as $z_{\text{prb}}$ should be influential to $z_{\text{prb}}$. In our example case, this means we expect the model to learn that the long training instances from class $1$ are positively influential for labeling a long probing example with class $1$. 
At the same time, the spurious attribute of $z_{\text{trn}}$ should not dominate the contribution to the training example's influence towards $z_{\text{prb}}$. Specifically, two long training examples, one with a confounding prefix \texttt{A} and the other with \texttt{B}, should be both influential to the long probing example with a confounding prefix, say, \texttt{B}.

\begin{figure}[t]
    \centering
    \includegraphics[width=0.45\textwidth]{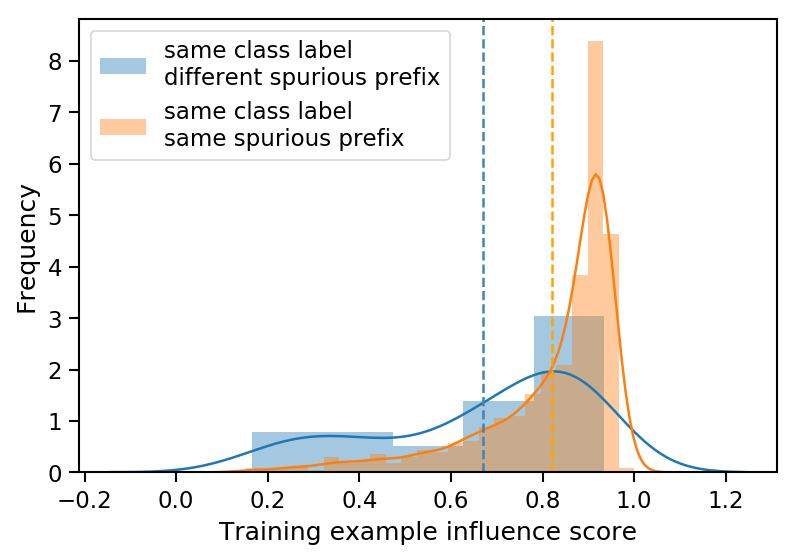}
    \caption{Distribution of each same-class training example's influence score $\mathcal{I}(z_{\text{trn}}, z_{\text{prb}})$ towards a typical probing example in TextLen (\Sref{sec:setup}). The range of influence scores is $[-1, 1]$. The average score difference between the two groups is $0.15$, and the difference is statistically significant via $t$-test.
    }
    \label{fig:illustrating}
\end{figure}

\autoref{fig:illustrating} illustrates the influence score distribution for a typical probing example. 
The probing example is from class $1$ (long text) and has a confounding prefix \texttt{B}. The orange plot in the figure shows the influence distribution of all class $1$ training examples with the same prefix \texttt{B}, whereas the blue plot shows the influence distribution of all class $1$ training examples with the different spurious prefix \texttt{A}. We observe that there is a statistically significant influence difference between these two groups. However, as the spurious attribute should not influence the model's decision process, we conjecture that \emph{the influence difference shows the model is confounded}.

\begin{figure*}[t]
    \centering
    \includegraphics[width=\textwidth]{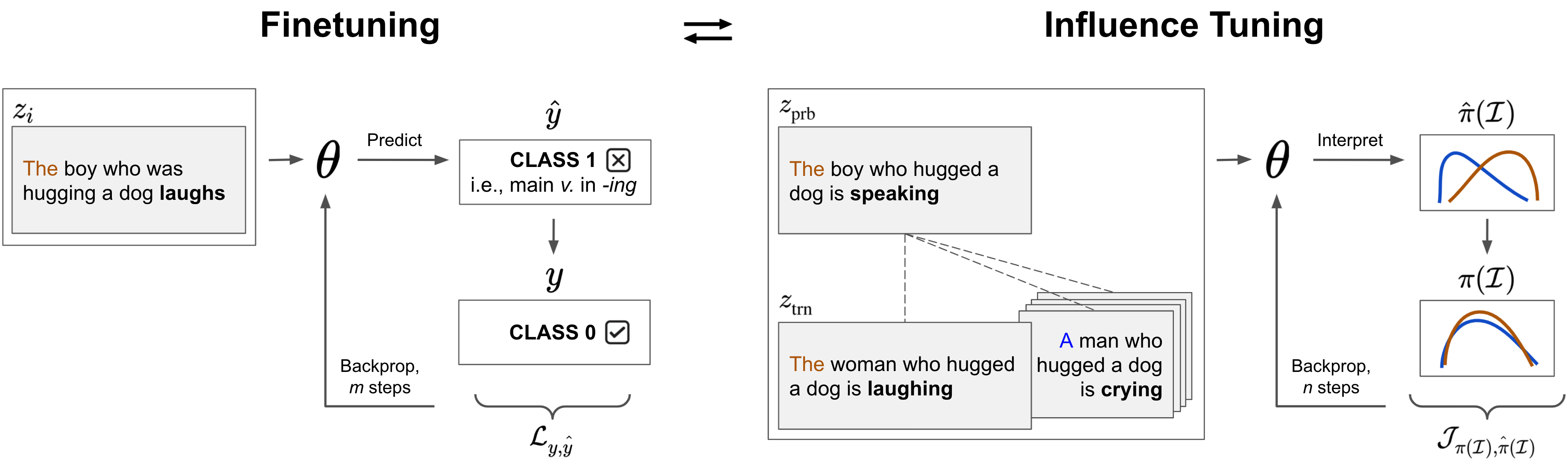}
    \caption{
    \footnotesize The influence tuning framework alternates between the \textbf{standard finetuning} phase (left) and the \textbf{influence tuning} phase (right). Illustrative examples are adapted from the MSGS dataset. 
    \textbf{Standard finetuning} (left): As introduced in \Sref{sec:setup}, the main verb in ``the boy who was hugging a dog \emph{laughs}'' is not a present participle ending with ``\emph{-ing}'', so the sentence should belong to \texttt{CLASS\_0} ($y$). A model $\theta$ may initially predict \texttt{CLASS\_1} with a high probability ($\hat{y}$). In the finetuning steps, we form a loss function $\mathcal{L}_{y, \hat{y}}$ over the labels and backpropagate into the model parameters.
    \textbf{Influence tuning} (right): For the influence tuning steps, we sample a few probing examples $z_{\text{prb}}$ and training examples $z_{\text{trn}}$ from the train set. In the figure $z_{\text{prb}}$ and $z_{\text{trn}}$ both belong to \texttt{CLASS\_1} (main verb in \emph{-ing} form) while the examples in $z_{\text{trn}}$ can have the same spurious attribute (sentence beginning with ``\emph{the}'') or different spurious attribute (beginning with ``\emph{a}'') as $z_{\text{prb}}$. A model $\theta$ may initially give the interpretation ($\hat{\pi}(\mathcal{I})$) that these examples in $z_{\text{trn}}$ have significantly different influence over $z_{\text{prb}}$. This could be a sign that the model is confounded; we form a loss function $\mathcal{J}_{\pi(\mathcal{I}), \hat{\pi}(\mathcal{I})}$ and backpropagate into the model parameters for a more plausible interpretation.
    }
    \label{fig:main_method}
\end{figure*}

As we show in this motivating example, 
research on interpretability via instance attribution 
can help us extract rationales behind the model decisions.
When we know what are possible spurious patterns in the data, we can check whether the spurious confounds are influencing learning, yielding implausible interpretations. 
For \emph{plausible} interpretations, such reliance on spurious attributes should not be as significant. 
In the next section, we propose a methodology to improve the model systematically, upon seeing an implausible rationale.

\section{Influence Tuning}
\label{sec:influence_tuning}

We propose a 
method to \emph{tune the model towards providing plausible rationales behind its decisions}. 
Motivated by the example scenario in \Sref{sec:motivation},
we define this plausibility to be the difference between the influence of training examples with different spurious attributes. 
We therefore call the method \emph{influence tuning}. 
Formally, we first randomly sample one probing example $z_{\text{prb}} = (x_{\text{prb}}, y_{\text{prb}})$ from the training set. 
We then sample a small group of training examples $\{z_{A_1}, \ldots, z_{A_k}\} \subset \{z_{\text{trn}}~|~y_{\text{trn}} = y_{\text{prb}}, c_{\text{trn}} = c_{\text{prb}}\}$, 
that share the same label ($y$) and \emph{the same} spurious attribute ($c$) as in $z_{\text{prb}}$ (e.g., samples from the orange distribution in \autoref{fig:illustrating}).
Similarly, we sample a small group of training examples $\{z_{B_1}, \ldots, z_{B_k}\} \subset \{z_{\text{trn}}~|~y_{\text{trn}} = y_{\text{prb}}, c_{\text{trn}} \neq c_{\text{prb}}\}$, that share the same label but with a spurious attribute that is \emph{different} from the spurious attribute in $z_{\text{prb}}$ (e.g., samples from the blue distribution in \autoref{fig:illustrating}). 
Note that although in our example scenario $y$ and $c$ are both binary, they are not required to be so. Since the spurious attribute $c$ should not be a part of the rationale behind the model's decision, we expect the average influence of $\{z_{A_1}, \ldots, z_{A_k}\}$ and $\{z_{B_1}, \ldots, z_{B_k}\}$ on $z_{\text{prb}}$ to be close to each other. Therefore, it is natural to define a new loss function over the influence scores and incorporate it in the model training:
\begin{align*}
    \mathcal{J} = (\frac{1}{k} \sum_{i=1}^k \mathcal{I}(z_{A_i}, z_{\text{prb}}) - \frac{1}{k} \sum_{i=1}^k \mathcal{I}(z_{B_i}, z_{\text{prb}}))^2 \text{.}
\end{align*}
To optimize for the influence loss $\mathcal{J}$, we need the gradient $\nabla_\theta \mathcal{J}$:
\begin{align*}
    &\nabla_\theta \mathcal{J} = (\frac{2}{k} \sum_{i=1}^k \mathcal{I}(z_{A_i}, z_{\text{prb}}) - \frac{2}{k} \sum_{i=1}^k \mathcal{I}(z_{B_i}, z_{\text{prb}}))\\
    &~~~(\frac{1}{k} \sum_{i=1}^k \nabla_\theta \mathcal{I}(z_{A_i}, z_{\text{prb}}) - \frac{1}{k} \sum_{i=1}^k \nabla_\theta \mathcal{I}(z_{B_i}, z_{\text{prb}})) \text{,}
\end{align*}
where the key is in calculating $\nabla_\theta \mathcal{I}(z_{\text{trn}}, z_{\text{prb}})$ with arbitrary $z_{\text{trn}}$ being either $z_{A_i}$ or $z_{B_i}$.

Recall that with the gradient cosine influence definition,  $\mathcal{I}(z_{\text{trn}}, z_{\text{prb}}) = \frac{\nabla_\theta \mathcal{L}(z_{\text{trn}}, \theta) \cdot \nabla_\theta \mathcal{L}(z_{\text{prb}}, \theta)}{\|\nabla_\theta \mathcal{L}(z_{\text{trn}}, \theta)\|~\|\nabla_\theta \mathcal{L}(z_{\text{prb}}, \theta)\|}$. We can then derive $\nabla_\theta \mathcal{I}(z_{\text{trn}}, z_{\text{prb}})$ as:
\begin{align*}
    &\nabla_\theta \mathcal{I}(z_{\text{trn}}, z_{\text{prb}}) = \mathbf{p} + \mathbf{q} - \mathbf{r} - \mathbf{s}\\
    &\mathbf{p} = \frac{1}{\|\nabla_\theta \mathcal{L}(z_{\text{trn}})\|~\|\nabla_\theta \mathcal{L}(z_{\text{prb}})\|} H_{\text{trn}} \nabla_\theta \mathcal{L}(z_{\text{prb}})\\
    &\mathbf{q} = \frac{1}{\|\nabla_\theta \mathcal{L}(z_{\text{trn}})\|~\|\nabla_\theta \mathcal{L}(z_{\text{prb}})\|} H_{\text{prb}} \nabla_\theta \mathcal{L}(z_{\text{trn}})\\
    &\mathbf{r} = \frac{\nabla_\theta \mathcal{L}(z_{\text{trn}}) \cdot \nabla_\theta \mathcal{L}(z_{\text{prb}})}{\|\nabla_\theta \mathcal{L}(z_{\text{trn}})\|^3~\|\nabla_\theta \mathcal{L}(z_{\text{prb}})\|} H_{\text{trn}} \nabla_\theta \mathcal{L}(z_{\text{trn}})\\
    &\mathbf{s} = \frac{\nabla_\theta \mathcal{L}(z_{\text{trn}}) \cdot \nabla_\theta \mathcal{L}(z_{\text{prb}})}{\|\nabla_\theta \mathcal{L}(z_{\text{trn}})\|~\|\nabla_\theta \mathcal{L}(z_{\text{prb}})\|^3} H_{\text{prb}} \nabla_\theta \mathcal{L}(z_{\text{prb}})
\end{align*}
where the Hessians $H_{\text{trn}} = \nabla_\theta^2 \mathcal{L}(z_{\text{trn}})$ and $H_{\text{prb}} = \nabla_\theta^2 \mathcal{L}(z_{\text{prb}})$. We omit $\theta$ in $\mathcal{L}(\cdot)$ for simplicity. Detailed derivations can be found in the appendix.\footnote{Intuitively, $\mathbf{p}$ and $\mathbf{q}$ find gradients that would help maximize the inner product of the training and probing model gradients in the next model update; $\mathbf{r}$ and $\mathbf{s}$ find gradients that would constrain the norm of the training and probing gradients for the next update.}

Overall, obtaining the gradient $\nabla_\theta \mathcal{J}$ for the influence loss $\mathcal{J}$ defined over the tuple $(z_{\text{prb}}, \{z_{A_1}, \ldots, z_{A_k}\}, \{z_{B_1}, \ldots, z_{B_k}\})$ makes the optimization possible. For the actual training process, we alternate the optimization of $\theta$ over both the regular label prediction loss $\mathcal{L}$ and the influence loss $\mathcal{J}$, with the interval as a hyperparameter to select.
That is, do $m$ steps of regular label loss propagation, $n$ steps of influence loss propagation, then back to label loss propagation, and so on. 
The goal is to find a set of model parameters, without changing the model architecture, that lead to both accurate label predictions and plausible rationales behind the decisions. 
We use a pretrained BERT model as our initial model $\theta$. 
\autoref{fig:main_method} summarizes our proposed method using the data examples that we will introduce in \Sref{sec:setup}.\footnote{Instead of the alternating optimization we adopted, folding the influence loss into the standard finetuning loss as a regularizer may work as well. We did not explore it here since our initial hypothesis is whether we can use a plausible interpretation to help build a more generalizable model: the instance attribution interpretation methods assume some regular, untouched finetuning steps before interpreting.}

\subsection{A special case of influence tuning}
\label{sec:special_case}
The above section gives an influence tuning framework based on the influence score $\mathcal{I}(\cdot, \cdot)$ defined on the full set of model parameters $\theta$. Now we are going to investigate an interesting special case of the framework, which defines the influence score on a partial parameter set.

Recall that we are using a pretrained BERT model as our initial model $\theta$ in our setup, and finetuning the BERT model would require training a prediction head over the transformer layers. For text classification, the prediction head is just a linear projection layer $W$, projecting from the BERT \texttt{[CLS]} token embedding to the label space and connecting to the final cross entropy loss. Additionally in our setup, our sampled $z_{\text{prb}}$ and $z_{\text{trn}}$ have the same label $y$. Now let's define a small parameter subset $\theta_{\text{proj}} = W_{(y)}$, representing the row of the final projection layer $W$ corresponding to the label $y$.

Similar to the original gradient cosine influence definition, we define $\mathcal{I}_{\text{proj}}(z_{\text{trn}}, z_{\text{prb}}) = \frac{\nabla_{\theta_{\text{proj}}} \mathcal{L}(z_{\text{trn}}, \theta) \cdot \nabla_{\theta_{\text{proj}}} \mathcal{L}(z_{\text{prb}}, \theta)}{\|\nabla_{\theta_{\text{proj}}} \mathcal{L}(z_{\text{trn}}, \theta)\|~\|\nabla_{\theta_{\text{proj}}} \mathcal{L}(z_{\text{prb}}, \theta)\|}$. We can further expand the label loss $\mathcal{L}$ with the parameter subset $W_{(y)}$ and the \texttt{[CLS]} embedding $h_{\texttt{[CLS]}}$:
\begin{align*}
    \mathcal{L}(z, \theta) =& -\log \frac{\exp(W_{(y)} h_{\texttt{[CLS]}})}{\sum_{y'} \exp(W_{(y')} h_{\texttt{[CLS]}})}\\
    \nabla_{\theta_{\text{proj}}} \mathcal{L}(z, \theta) =& \frac{\exp(W_{(y)} h_{\texttt{[CLS]}})}{\sum_{y'} \exp(W_{(y')} h_{\texttt{[CLS]}})} h_{\texttt{[CLS]}}\\
    &- h_{\texttt{[CLS]}} = (p(y) - 1) h_{\texttt{[CLS]}}
\end{align*}
Therefore, $\frac{\nabla_{\theta_{\text{proj}}} \mathcal{L}(z_{\text{trn}}, \theta)}{\|\nabla_{\theta_{\text{proj}}} \mathcal{L}(z_{\text{trn}}, \theta)\|} = -\frac{h_{\text{trn}\texttt{[CLS]}}}{\|h_{\text{trn}\texttt{[CLS]}}\|}$, and similarly $\frac{\nabla_{\theta_{\text{proj}}} \mathcal{L}(z_{\text{prb}}, \theta)}{\|\nabla_{\theta_{\text{proj}}} \mathcal{L}(z_{\text{prb}}, \theta)\|} = -\frac{h_{\text{prb}\texttt{[CLS]}}}{\|h_{\text{prb}\texttt{[CLS]}}\|}$. We finally rewrite the partial influence definition $\mathcal{I}_{\text{proj}}(z_{\text{trn}}, z_{\text{prb}})$ as $\frac{h_{\text{trn}\texttt{[CLS]}} \cdot h_{\text{prb}\texttt{[CLS]}}}{\|h_{\text{trn}\texttt{[CLS]}}\|~\|h_{\text{prb}\texttt{[CLS]}}\|}$, \emph{which is essentially the cosine similarity between the training and probing example's \texttt{[CLS]} embeddings}.

The new definition $\mathcal{I}_{\text{proj}}(z_{\text{trn}}, z_{\text{prb}})$ leads to a new influence loss $\mathcal{J}_{\text{proj}}$. Different from the second-order influence tuning method that obtains $\nabla_\theta \mathcal{J}$, we can get $\nabla_\theta \mathcal{J}_{\text{proj}}$ by applying the regular gradient backward operation on the model and thus updating the model faster. All the other parts of the framework, like the data tuple selection and the alternating training objectives, remain the same. We call this special variant of influence tuning \textbf{embedding tuning}.

\section{Experimental Setup}
\label{sec:experimental_setup}
\subsection{Adversarial training as a baseline}
\label{sec:adv_train} 
One notable feature of influence tuning is that it is designed to help deconfounding NLP models without adding additional modules to the network. A related line of research on deconfounding NLP models takes the intuition from domain adversarial training \citep{ganin2016domain, pryzant2018deconfounded, kumar2019topics}. These methods usually have two classifier modules built on top of a shared encoder. The objective for the model is adversarial: the model should be able to predict the target label $y$ of the input accurately, while failing at reconstructing the spurious attribute $c$ effectively, which potentially indicates that the confounding information regarding $c$ is not encoded by the model.

As a baseline for this work, we specifically modify a BERT model according to the method described in \citet{pryzant2018deconfounded}. It uses a gradient reversal layer at the beginning of the confound classifier head that would multiply the gradient by $-1$ during the backward pass. All of the BERT transformer layers form the shared encoder for the label classifier and the confound classifier. The implicit loss used by the model can then by written as $\mathcal{L} = \mathcal{L}_{\text{label}} - \lambda \mathcal{L}_{\text{confound}}$, where $\mathcal{L}_{\text{label}}$ and $\mathcal{L}_{\text{confound}}$ are both cross entropy loss, and $\lambda$ is a hyperparameter to select. 
Essentially, this method learns and unlearns---learns to predict the correct label while unlearning the information that could help reconstruct the confound attribute. 
Compared to other prior work tackling spurious correlations mentioned in \Sref{sec:intro}, this method is also most suitable for a direct comparison with our proposed influence tuning framework, because both methods aim to explicitly demote certain known confounds for the model.

\subsection{Datasets}
\label{sec:setup}
To evaluate the proposed approaches for deconfounding NLP models, we conduct controlled experiments on two datasets.

\paragraph{TextLen}
TextLen is a synthetic dataset we create that follows the example scenario in \Sref{sec:motivation}. Specifically, for the training set, we randomly split 1500 sentences from the canonical CoNLL-2003 shared task dataset \citep{sang2003introduction} to two classes $0$ and $1$ of equal sizes. Sentences from class $0$ are trimmed to a length sampled from a normal distribution with $\mu_{\text{short}} = 15, \sigma = 4$; sentences from class $1$ are trimmed with $\mu_{\text{long}} = 25, \sigma = 4$. We add prefix tokens \texttt{A}$=$``Negative.'' and \texttt{B}$=$``Positive.'' to the start of the sentences, such that $90\%$ of the time, a class $0$ sentence would receive the negative prefix and a class $1$ sentence would receive the positive prefix. However, in the dev set and the test set of TextLen,\footnote{Like the TextLen training set, the dev set content also comes from the CoNLL-2003 training set; the TextLen test set content instead comes from the CoNLL-2003 test set.} while trimmed with the same text length distribution, only $50\%$ of the time the confounding prefix would correlate with the class label of the sentence. A deconfounded model should achieve a good classification performance on both the train and test splits.

\paragraph{MSGS} The \textbf{M}ixed \textbf{S}ignals \textbf{G}eneralization \textbf{S}et is proposed by \citet{warstadt2020learning} to investigate whether language models would acquire a preference for linguistic generalizations. The model is supposed to learn a classification task with an ambiguous training dataset. For example, a class $1$ sentence could be ``\emph{the boy who hugged a cat is sneezing}'', and a class $0$ sentence could be ``\emph{a boy who is hugging the cat sneezed}''. To distinguish the two classes, a model performing surface generalizations could potentially rely on whether the article ``\emph{the}'' or ``\emph{a}'' precedes the sentence. A model performing linguistic generalizations, however, could be deciding based on whether the main verb of the sentence is in the ``\emph{-ing}'' form. The linguistic feature decides the class of the sentence in both the MSGS train and test sets, whereas the surface feature correlates highly with the classes only in the training set, and co-occurs randomly with the classes in the disambiguated test data. Specifically, we choose MSGS's \textsc{syntactic category} as the core attribute and \textsc{relative position} as the spurious attribute; for the training set, we also adopt an inoculation rate of $0.3\%$ \citep{warstadt2020learning}.

We show statistics of TextLen and MSGS in \autoref{tab:dataset}. We use BERT-base as our model template and the default BERT Adam optimizer for both tasks and all of the deconfounding methods. We perform hyperparameter search using the dev set for all of the methods. Detailed hyperparameters can be found in the appendix.
\begin{table*}[ht]
    \centering
    \begin{tabular}{@{}lp{1in}p{1.5in}@{}}
    \toprule
     & \textbf{TextLen} & \textbf{MSGS}\\
      \midrule
      Size & 1500 / 480 / 500 & 5000 / 15000 / 15000\\
      Core attribute & Text length & \textsc{syntactic category}\\
      Spurious attribute & Prefix token & \textsc{relative position}\\
      Variance in core attribute & Yes & No\\
      Spurious attribute train set success rate & $90.0\%$ & $99.7\%$\\
      Spurious attribute test set success rate & $50.0\%$ & $66.7\%$\\
      \bottomrule
    \end{tabular}
    \caption{Dataset statistics}
    \label{tab:dataset}
\end{table*}

\section{Results}
\label{sec:results}
\subsection{Does influence tuning make the model interpretations more plausible?}
We are interested in a preliminary research question first: having seen the confounded model interpretations discovered in \Sref{sec:motivation}, does our proposed method, influence tuning, make the model interpretations more plausible? To quantitatively measure how much the model relies on the spurious attribute to make decisions, for both tasks we randomly select 40 probing examples $z_{\text{prb}}$ from the training set. For each probing example $z_{\text{prb}}$, we put the training examples into two groups: $A = \{z_{\text{trn}}~|~y_{\text{trn}} = y_{\text{prb}}, c_{\text{trn}} = c_{\text{prb}}\}$ and $B = \{z_{\text{trn}}~|~y_{\text{trn}} = y_{\text{prb}}, c_{\text{trn}} \neq c_{\text{prb}}\}$, where $y$ is the true label and $c$ is the confounding spurious attribute. We define the \emph{confound influence difference} (CID) to be the influence difference between the two groups to the probing example: $\frac{1}{|A|} \sum_{\text{trn} \in A} \mathcal{I}(z_{\text{trn}}, z_{\text{prb}}) - \frac{1}{|B|} \sum_{\text{trn} \in B} \mathcal{I}(z_{\text{trn}}, z_{\text{prb}})$.
\begin{figure}[t]
    \centering
    \includegraphics[width=0.45\textwidth]{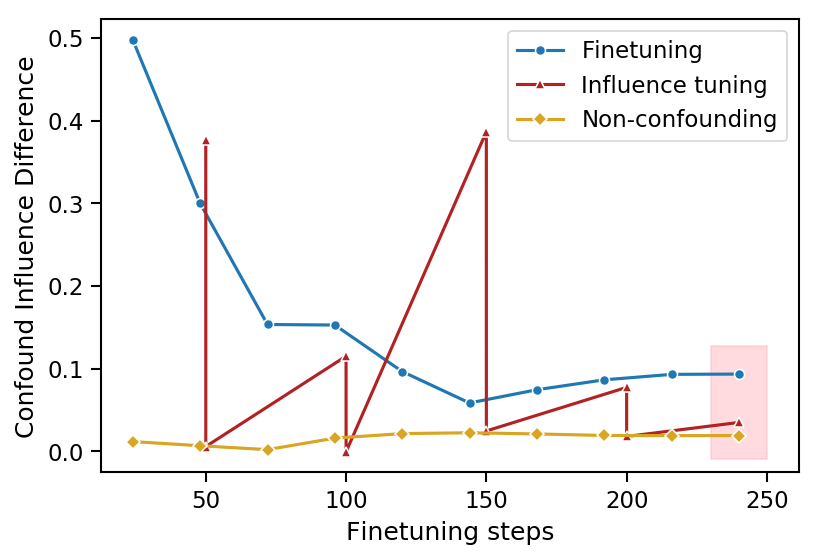}
    \caption{Average confound influence difference (CID) of different models in the TextLen task. The scale of the influence scores is $[-1, 1]$. The final CID is $0.093$ for the finetuning model, $0.035$ for the influence tuning model, and $0.019$ for the model trained on non-spurious data.
    }
    \label{fig:cid}
\end{figure}

We show in \autoref{fig:cid} the average CID of three different models for TextLen during the training process: a model that is trained with the \emph{regular finetuning} objective, a model that is trained using the \emph{influence tuning} framework, and a control model that is trained over a \emph{non-confounding} version of TextLen data (i.e., the spurious prefix token is removed). The final CIDs are $0.093$, $0.035$ and $0.019$, respectively for the three models. We observe that both the finetuning model and the influence tuning model start with a very high CID, indicating the confound attribute is exploited heavily at the beginning of the training process.
However, for the influence tuning model, each influence tuning round happened after every 50 standard finetuning steps, helps the model achieve a near zero CID (as shown by the vertical drops within the influence tuning plot). The model does regain the CID during the following finetuning steps, but eventually arrives at a relatively low CID. The result on the MSGS data is similar, except that we do not have the non-confounding control model. The drops in CID caused by influence tuning answer our preliminary question: we do find that influence tuning makes the model interpretations more plausible in accordance with our expectation.

\subsection{Does the guided plausibility transform well to the model generalizability?}
\label{sec:plausible_to_generalizable}

Would a more plausible model, or more specifically a model that is \emph{guided} to provide plausible interpretations, achieve a stronger performance in 
out-of-distribution 
test sets, where the confound information no longer helps the decision? 
To answer this question, we compare the different deconfounding approaches introduced in \Sref{sec:influence_tuning} and \Sref{sec:adv_train} over the TextLen and MSGS tasks.
\begin{table*}[t]
    \centering
    \begin{tabular}{@{}llllll@{}}
    \toprule
    & Finetuning & Adversarial training & Influence tuning & Embedding tuning & No spurious \\
      \midrule
      \textbf{TextLen} & $76.00$ & $78.44_{~(p=0.026)}$ & $80.48_{~(p=0.003)}$ & $\mathbf{82.20}_{~(p<0.001)}$ & $84.96_{~(p<0.001)}$\\
      \textbf{MSGS} & $75.33$ & $83.41_{~(p=0.179)}$ & $\mathbf{88.23}_{~(p=0.015)}$ & $87.45_{~(p=0.042)}$ & -\\
      \bottomrule
    \end{tabular}
    \caption{Test set performance (accuracy) of different deconfounding approaches. All the experiments use five random seeds. The subscripts show the $p$-values of the $t$-tests comparing the deconfounding approaches with the regular finetuning model.}
    \label{tab:test_set_performance}
\end{table*}

We show our main results in \autoref{tab:test_set_performance}. On TextLen, we observe that the adversarial training method 
gives a moderate improvement over the regular finetuning model. 
Both influence tuning and embedding tuning lead to significant accuracy gain, with the embedding tuning method achieves the highest 82.2\% test accuracy. In \Sref{sec:special_case} we derived why embedding tuning is a special case of influence tuning with a reduced set of parameters for the influence calculation. This could be enough for the TextLen dataset since the task is relatively easy. The upper bound model with the spurious attribute removed from the data still outperforms all of the deconfounding methods, leaving a gap to address in future work.

On MSGS, we observe a similar trend for the adversarial training method, 
which makes a moderate improvement over finetuning as expected.
Again, both influence tuning and embedding tuning achieve significant improvements. However, the influence tuning method outperforms embedding tuning on this task. One contributing reason could be that the linguistic generalizations required by this task can be encoded across the full BERT transformer layers. Therefore, the influence defined only over the final projection layer in the embedding tuning case might be limiting. Overall, both our proposed influence tuning and the special case embedding tuning are effective at deconfounding the models in our experiments compared to the baselines.

\subsection{Can we use less data with influence tuning?}
The advantage of influence tuning does not come without a price. For a general dataset, it would require at least some lightweight annotations in addition to the regular label information. For example, in \Sref{sec:influence_tuning} when we operate with the data tuple $(z_{\text{prb}}, \{z_{A_1}, \ldots, z_{A_k}\}, \{z_{B_1}, \ldots, z_{B_k}\})$, we would need information about the confound group an example belongs to. Though in our experiments with TextLen and MSGS we are sampling a relatively small set of $z_{\text{prb}}$, $z_{A_i}$ and $z_{B_i}$ (50--100 $z_{\text{prb}}$ each full influence tuning round, 3--5 $z_{A_i}$ and $z_{B_i}$ for each $z_{\text{prb}}$), the model still potentially has access to the confound information of the full dataset during the whole training process.\footnote{Our baseline approach adversarial training also has access to and actually uses the confound information of the full data.} Therefore, in this section we are interested in whether we can strictly provide less accessible confound information to the model, and how this would affect the performance of our methods.

For both the TextLen and MSGS data, we randomly select subsets of the training set containing $m$\% of the total examples. Then during the training process, we limit the model to sample $z_{\text{prb}}$, $z_{A_i}$ and $z_{B_i}$ only from the $m$\% training subset where the confound group information is accessible. Note that this serves as a hard \emph{upper limit} for the confound access, and the actual confound information queried by the influence tuning framework can be well under this limit.

In \autoref{tab:data_ablation} we show the test performance of influence tuning and embedding tuning on TextLen and MSGS, using the same model hyperparameters as the results in \autoref{tab:test_set_performance}. However, unlike the \autoref{tab:test_set_performance} results where every trial within the five random seeds succeeds in fitting the \emph{training set}, the experiments with the data constraint sometimes fail to converge (i.e., not even fitting the train set). We exclude such failed trials from the average performance reported in \autoref{tab:data_ablation}, while we observe that at least three out of the five trials for each confound access rate can converge successfully.\footnote{Importantly, the trials are considered failed based on their training set performance (0.5 accuracy equivalent to random guess), not based on the dev set or test set performance. When deploying to an unknown test set, we would know when to re-train the model based on its known training performance.} We see that even with the hard constraint on the confound access rate, influence tuning and embedding tuning still outperform the baseline methods, using only 5\%-20\% examples. Generally, higher confound access rate would lead to stronger deconfounding performance, which creates a tradeoff to decide based on the need of the users.
\begin{table}[t]
    \centering
    \begin{tabular}{@{}lp{1in}p{1.1in}@{}}
    \toprule
     & Influence tuning & Embedding tuning\\
      \midrule
      \textbf{TextLen}\\
      5\% & $78.24$ & $78.84$\\
      10\% & $78.68$ & $80.80$\\
      20\% & $80.65$ & $80.44$\\
      50\% & $80.13$ & $81.07$\\
      100\% & $80.48$ & $82.20$\\
      [4pt]
      \textbf{MSGS}\\
      5\% & $81.74$ & $84.90$\\
      10\% & $80.48$ & $81.56$\\
      20\% & $83.90$ & $81.56$\\
      50\% & $93.57$ & $82.80$\\
      100\% & $88.23$ & $87.45$\\
      \bottomrule
    \end{tabular}
    \caption{Performance of influence tuning and embedding tuning when there is an upper limit for the confound access rate. The accuracy shown is an average of at least three succeeded trials within the use of five random seeds.}
    \label{tab:data_ablation}
\end{table}

\subsection{Discussion}
\label{sec:discussion}
In this section we answered three questions regarding the influence tuning framework: whether it makes the model more plausible, whether it helps the model achieve a strong deconfounding performance, and whether it can be used with a reduced amount of data. We conducted experiments on a synthetic dataset and a linguistic probing dataset, but the potential application of our approach can be more impactful than the current tasks. For example, our method might be helpful for identifying and mitigating gender biases and racial biases in sentiment analysis or toxicity detection systems \citep{kiritchenko2018examining,Sap2019TheRO}, by modeling the problem as a deconfounding task. 
One potential drawback is that these natural cases would inevitably require some extent of extra human annotations, over the known or hypothesized confounds in the problem. 
However, we also believe the human feedback in NLP \citep{Settles2011ClosingTL,Kaushik2020LearningTD,Wang2021PuttingHI} is a crucial and controllable way to tackle model’s exploitation of spurious correlations in the data, which happens as a result of the absence of proper supervision.
Furthermore, if we define the influence objective differently in \Sref{sec:influence_tuning}, e.g. modeling which groups of examples \emph{should} be influential to the probing instance and to what extent, we may be able to implicitly \emph{promote} the core attributes in the task in addition to demoting the confounds.

\section{Related Work}
Interpreting NLP models by token-level importance scores over the input span is a widely adopted approach \citep{belinkov2019analysis}. These scores can be gradient-based \citep{Simonyan2013DeepIC, Li2016VisualizingAU}, attention weights if supported by the model \citep{Jain2019AttentionIN, Wiegreffe2019AttentionIN}, or weights from a linear model fitting the local region of a deep model \citep{Ribeiro2016WhySI}. The models can achieve better performance or learn more efficiently if supervisions are provided for these feature importance scores \citep{Ross2017RightFT,zhong2019fine,Pruthi2020EvaluatingEH}.

Unlike the token-level interpretations, our focus in this work is on the instance attribution methods. Apart from influence functions \citep{Koh2017UnderstandingBP} and TracIn \citep{Pruthi2020EstimatingTD} that are already introduced, other instance attribution methods include representer point selection \citep{Yeh2018RepresenterPS} and $\theta$-relative influence functions \citep{Barshan2020RelatIFIE}, with \citet{pezeshkpour2021empirical} comparing the methods empirically in NLP tasks. 
However, these methods do not facilitate a systematic improvement for the model based on the plausibility of the interpretations, which is a gap addressed by this work. 
Models designed with explicit interpretability considerations like deep weighted averaging classifiers \citep{Card2019DeepWA} and SelfExplain \citep{rajagopal2021selfexplain} may also support instance attribution, though the flexibility of the model architecture can be more limited. 
\citet{charpiat2019input} also compute and enforce a gradient-based cosine similarity between examples in computer vision tasks. The intuition is that similar examples should be seen as similar by the model, and they observe a faster convergence in model training.

One key use case of the proposed influence tuning framework is to deconfound the model from relying on spurious attributes during the decision process. Other works that aim at preventing neural models from using the spurious attributes include \citet{elazar2018adversarial} and \citet{pryzant2018deconfounded} which operate over a known set of confounds, and \citet{kumar2019topics} which models unknown, latent confounds. They often involve the idea of learning invariant features across domains through adversarial training \citep{ganin2016domain, xie2017controllable}. Spurious correlations can also be mitigated by data-based, optimization-based, and post-processing methods \citep{zmigrod-etal-2019-counterfactual,Kaushik2020LearningTD,Sagawa2019DistributionallyRN,Yaghoobzadeh2021IncreasingRT,clark2019don}. 
In this work, we mainly compare with the adversarial training method with gradient reversal in \citet{pryzant2018deconfounded} as a baseline, since both methods perform explicit demotions to some known confounds in the data used by the model. 
Future work can explore comparisons and potential combinations with other approaches addressing the spurious correlations.

\section{Conclusion}
NLP models that build upon deep neural networks are notoriously opaque about their decision process. Though instance attribution methods can be used to unveil problems of the model reflected by the implausible interpretations, a novel research question is whether or how the model training can benefit from interpretability methods in a systematic way. Our work addresses this question, by proposing the influence tuning framework that backpropagates a target instance attribution interpretation directly to the model. 
In two use cases of demoting spurious confounds in the data, we show that (1) influence tuning can eventually lead to more plausible model interpretations; (2) influence tuning can help build better-performing deconfounded models compared to those trained with the baseline methods; (3) influence tuning can still be reliable in lower-resource setups. 
Future work will explore more datasets and tasks, and other optimization methods. Additionally, we will explore guiding the model to learn to \emph{promote} core attributes of the task in addition to \emph{demoting} the spurious confounds.

\section*{Acknowledgments}
We thank Haoping Bai, Sachin Kumar, and the anonymous EMNLP reviewers and area chairs for helpful discussions of this work. 
This material is based upon work funded by the DARPA CMO under Contract No.~HR001120C0124. The views and opinions of authors expressed herein do not necessarily state or reflect those of the United States Government or any agency thereof.

\bibliography{my_cites}
\bibliographystyle{acl_natbib}

\appendix

\section{Hyperparameters}
For finetuning BERT, we use a learning rate of 2e-5 for both TextLen and MSGS. We tune 10 epochs for TextLen and 3 epochs for MSGS with a batch size of 64, resulting in around 250 steps for each dataset since the data size is also different.

For adversarial training, we use the same batch size of 64, but search the learning rate $\in \{\text{2e-5, 5e-5}\}$, the number of training epochs $\in \{\text{10, 20, 40}\}$ for TextLen and $\in \{\text{3, 6, 12}\}$ for MSGS, and $\lambda \in \{\text{0.1, 0.3, 1.0, 3.0}\}$. For TextLen, the best hyperparameters are [5e-5, 20, 0.3]. For MSGS, the best hyperparameters are [5e-5, 12, 0.3].

For influence tuning and embedding tuning, we follow the vanilla finetuning and use the same learning rate of 2e-5, batch size of 64 and a total number of steps at around 250 for the regular label loss propagation steps. For the influence loss propagation steps, we search the tuning interval (i.e., how many label propagation steps needed before a round of influence propagation happens) $\in \{\text{25, 50, 100}\}$, number of epochs within one round of influence propagation $\in \{\text{5, 10, 15}\}$ for TextLen and $\in \{\text{3}\}$ for MSGS, batch size $\in \{\text{4, 16, 64}\}$, influence propagation optimizer's learning rate $\in \{\text{3e-5, 1e-5, 3e-6, 1e-6}\}$. The best hyperparameters for influence tuning are [50, 5, 64, 3e-5] for TextLen and [50, 3, 16, 3e-5] for MSGS. The best hyperparameters for embedding tuning are [50, 10, 64, 3e-5] for TextLen and [25, 3, 4, 1e-5] for MSGS. For TextLen, each influence tuning epoch we randomly sample 75 probing examples ($z_{\text{prb}}$) from the train set, and for each probing example we sample 5 positive and 5 negative train examples based on the confound information ($z_{A_i}$ and $z_{B_i}$). For MSGS, each influence tuning epoch we randomly sample 100 probing examples ($z_{\text{prb}}$) from the train set, and for each probing example we sample 3 positive and 3 negative train examples based on the confound information ($z_{A_i}$ and $z_{B_i}$). For all experiments with all the methods, we use 5 random seeds [2021, 2022, 2023, 2024, 2025].

\section{Derivation of the influence gradients}
To derive the gradient of the cosine influence, we first derive the gradient of the dot product influence:
\footnotesize
\begin{align*}
    I =& \nabla_\theta L(x_1) \cdot \nabla_\theta L(x_2)\\
    =& \frac{\partial L(x_1)}{\partial \theta_1} \frac{\partial L(x_2)}{\partial \theta_1} + \frac{\partial L(x_1)}{\partial \theta_2} \frac{\partial L(x_2)}{\partial \theta_2} + \ldots
    \\
    \frac{\partial}{\partial \theta_1} I =& (\frac{\partial^2 L(x_1)}{\partial \theta_1^2} \frac{\partial L(x_2)}{\partial \theta_1} + \frac{\partial L(x_1)}{\partial \theta_1} \frac{\partial^2 L(x_2)}{\partial \theta_1^2})\\
    &+ (\frac{\partial^2 L(x_1)}{\partial \theta_1 \partial \theta_2} \frac{\partial L(x_2)}{\partial \theta_2} + \frac{\partial L(x_1)}{\partial \theta_2} \frac{\partial^2 L(x_2)}{\partial \theta_1 \partial \theta_2})\\
    &+ \ldots\\
    =& H_{L1}[\text{row 1}] \cdot \nabla_\theta L(x_2)\\
    &+ H_{L2}[\text{row 1}] \cdot \nabla_\theta L(x_1)
    \\
    \ldots&\\
    \nabla_\theta I =& H_{L1} \nabla_\theta L(x_2) + H_{L2} \nabla_\theta L(x_1)
\end{align*}
\normalsize

Next we derive the gradient of the full cosine influence:
\footnotesize
\begin{align*}
    I =& \frac{\nabla_\theta L(x_1) \cdot \nabla_\theta L(x_2)}{\|\nabla_\theta L(x_1)\|~\|\nabla_\theta L(x_2)\|} = \frac{i(\theta)}{m(\theta)}\\
    \nabla_\theta I =& \frac{m(\theta) \nabla_\theta i(\theta) - i(\theta) \nabla_\theta m(\theta)}{m(\theta)^2}\\
\end{align*}
\normalsize

We already know the gradient of $i(\theta)$, so we are only interested in $\nabla_\theta m(\theta)$. We first calculate $\nabla_\theta \|\nabla_\theta L(x_1)\|$ and $\nabla_\theta \|\nabla_\theta L(x_2)\|$, and then apply product rule to combine:
\footnotesize
\begin{align*}
    \frac{\partial}{\partial \theta_1}& \|\nabla_\theta L(x_1)\|\\
    =& \frac{\partial}{\partial \theta_1} ((\frac{\partial}{\partial \theta_1} L(x_1))^2 + (\frac{\partial}{\partial \theta_2} L(x_1))^2 + \ldots)^{1/2}\\
    =& \frac{1}{2~\|\nabla_\theta L(x_1)\|} (\frac{\partial}{\partial \theta_1}(\frac{\partial}{\partial \theta_1} L(x_1))^2 + \frac{\partial}{\partial \theta_1}(\frac{\partial}{\partial \theta_2} L(x_1))^2\\
    &\qquad \qquad \qquad + \ldots)\\
    =& \frac{1}{2~\|\nabla_\theta L(x_1)\|} (2\frac{\partial}{\partial \theta_1}L(x_1) \frac{\partial^2}{\partial \theta_1^2}L(x_1) +\\
    &\qquad \qquad \qquad \qquad 2\frac{\partial}{\partial \theta_2}L(x_1) \frac{\partial}{\partial \theta_1}\frac{\partial}{\partial \theta_2}L(x_1) + \ldots)\\
    =& \frac{1}{\|\nabla_\theta L(x_1)\|} (H_{L1}[\text{row 1}] \cdot \nabla_\theta L(x_1))
    \\
    \ldots\\
    \nabla_\theta& \|\nabla_\theta L(x_1)\|\\
    =& \frac{1}{\|\nabla_\theta L(x_1)\|} H_{L1} \nabla_\theta L(x_1)
    \\
    \nabla_\theta& \|\nabla_\theta L(x_2)\|\\
    =& \frac{1}{\|\nabla_\theta L(x_2)\|} H_{L2} \nabla_\theta L(x_2)
    \\
    \nabla_\theta& m(\theta)\\
    =& \frac{\|\nabla_\theta L(x_2)\|}{\|\nabla_\theta L(x_1)\|} H_{L1} \nabla_\theta L(x_1)\\
    &+ \frac{\|\nabla_\theta L(x_1)\|}{\|\nabla_\theta L(x_2)\|} H_{L2} \nabla_\theta L(x_2)
\end{align*}
\normalsize

Combining the above terms, we have:
\tiny
\begin{align*}
    \nabla_\theta I =& \frac{\|\nabla_\theta L(x_1)\|~\|\nabla_\theta L(x_2)\|(H_{L1} \nabla_\theta L(x_2) + H_{L2} \nabla_\theta L(x_1))}{\|\nabla_\theta L(x_1)\|^2~\|\nabla_\theta L(x_2)\|^2}\\
    & - \frac{(\nabla_\theta L(x_1) \cdot \nabla_\theta L(x_2))(\frac{\|\nabla_\theta L(x_2)\|}{\|\nabla_\theta L(x_1)\|} H_{L1} \nabla_\theta L(x_1))}{\|\nabla_\theta L(x_1)\|^2~\|\nabla_\theta L(x_2)\|^2}\\
    & - \frac{(\nabla_\theta L(x_1) \cdot \nabla_\theta L(x_2))(\frac{\|\nabla_\theta L(x_1)\|}{\|\nabla_\theta L(x_2)\|} H_{L2} \nabla_\theta L(x_2))}{\|\nabla_\theta L(x_1)\|^2~\|\nabla_\theta L(x_2)\|^2}
\end{align*}
\normalsize

For clarity, we rearrange the equation as below:
\begin{align*}
    \nabla_\theta I =& \mathbf{p} + \mathbf{q} - \mathbf{r} - \mathbf{s}\\
    \mathbf{p} =& \frac{1}{\|\nabla_\theta L(x_1)\|~\|\nabla_\theta L(x_2)\|} H_{L1} \nabla_\theta L(x_2)\\
    \mathbf{q} =& \frac{1}{\|\nabla_\theta L(x_1)\|~\|\nabla_\theta L(x_2)\|} H_{L2} \nabla_\theta L(x_1)\\
    \mathbf{r} =& \frac{\nabla_\theta L(x_1) \cdot \nabla_\theta L(x_2)}{\|\nabla_\theta L(x_1)\|^3~\|\nabla_\theta L(x_2)\|} H_{L1} \nabla_\theta L(x_1)\\
    \mathbf{s} =& \frac{\nabla_\theta L(x_1) \cdot \nabla_\theta L(x_2)}{\|\nabla_\theta L(x_1)\|~\|\nabla_\theta L(x_2)\|^3} H_{L2} \nabla_\theta L(x_2)
\end{align*}

\end{document}